\documentclass[journal]{IEEEtran}
\usepackage[numbers,sort&compress]{natbib}
\usepackage[colorlinks,linkcolor=green,anchorcolor=green,citecolor=blue]{hyperref}
\usepackage{amsmath,amssymb,amsfonts}
\usepackage{algorithmic}
\usepackage{color}
\usepackage{graphicx}
\usepackage[caption=false,font=normalsize,labelfont=sf,textfont=sf]{subfig}
\usepackage{textcomp}
\usepackage{xcolor}
\def\BibTeX{{\rm B\kern-.05em{\sc i\kern-.025em b}\kern-.08em
    T\kern-.1667em\lower.7ex\hbox{E}\kern-.125emX}}
\usepackage{ragged2e}
\usepackage{booktabs,makecell, multirow, tabularx}
\usepackage{url}

\begin{document}
\title{SDVRF: Sparse-to-Dense Voxel Region Fusion for Multi-modal 3D Object Detection}
\author{}
\author{
Binglu Ren,
Jianqin Yin* \IEEEmembership{Member, IEEE}
\thanks{* Corresponding author}
}

\maketitle

\begin{abstract}
In the perception task of autonomous driving, multi-modal methods have become a trend due to the complementary characteristics of LiDAR point clouds and image data. However, the performance of multi-modal methods is usually limited by the sparsity of the point cloud or the noise problem caused by the misalignment between LiDAR and the camera. To solve these two problems, we present a new concept, Voxel Region (VR), which is obtained by projecting the sparse local point clouds in each voxel dynamically. And we propose a novel fusion method named Sparse-to-Dense Voxel Region Fusion (SDVRF). Specifically, more pixels of the image feature map inside the VR are gathered to supplement the voxel feature extracted from sparse points and achieve denser fusion. Meanwhile, different from prior methods, which project the size-fixed grids, our strategy of generating dynamic regions achieves better alignment and avoids introducing too much background noise. Furthermore, we propose a multi-scale fusion framework to extract more contextual information and capture the features of objects of different sizes. Experiments on the KITTI dataset show that our method improves the performance of different baselines, especially on classes of small size, including Pedestrian and Cyclist. 
\end{abstract}

\begin{IEEEkeywords}
3D object detection, multi-modal, autonomous driving
\end{IEEEkeywords}

\section{Introduction}
3D object detection aims to predict objects' locations, sizes, and classes in the 3D space with sensory input, which is a fundamental and crucial task in an automotive perception system \cite{mao20223d}. Reliable observations of objects around the ego vehicle will facilitate downstream components like trajectory prediction and path planning to enhance road safety \cite{mao20223d, qian20223d}. Among the numerous onboard sensors, LiDAR has great penetrability and provides accurate geometric information. Moreover, the camera provides rich semantic information for detection \cite{wang2021multi}.

To leverage the complementary characteristics of different sensors, researchers have paid more and more attention to LiDAR-camera 3D object detection in recent years \cite{mao20223d}. However, there is a huge gap between the representations of images and point clouds. Because of the physical limitations of LiDAR, points become extremely sparse when objects are far from the ego or occluded. Unlike point clouds, images can provide dense pixels even when objects are far apart. But it is hard to get high-accuracy 3D bounding boxes because of the lack of depth information in images. Therefore, how to fuse the sparse points and the dense pixels is the key to LiDAR-camera fusion.

\begin{figure}[htp]
\centerline{\includegraphics[width=0.45\textwidth]{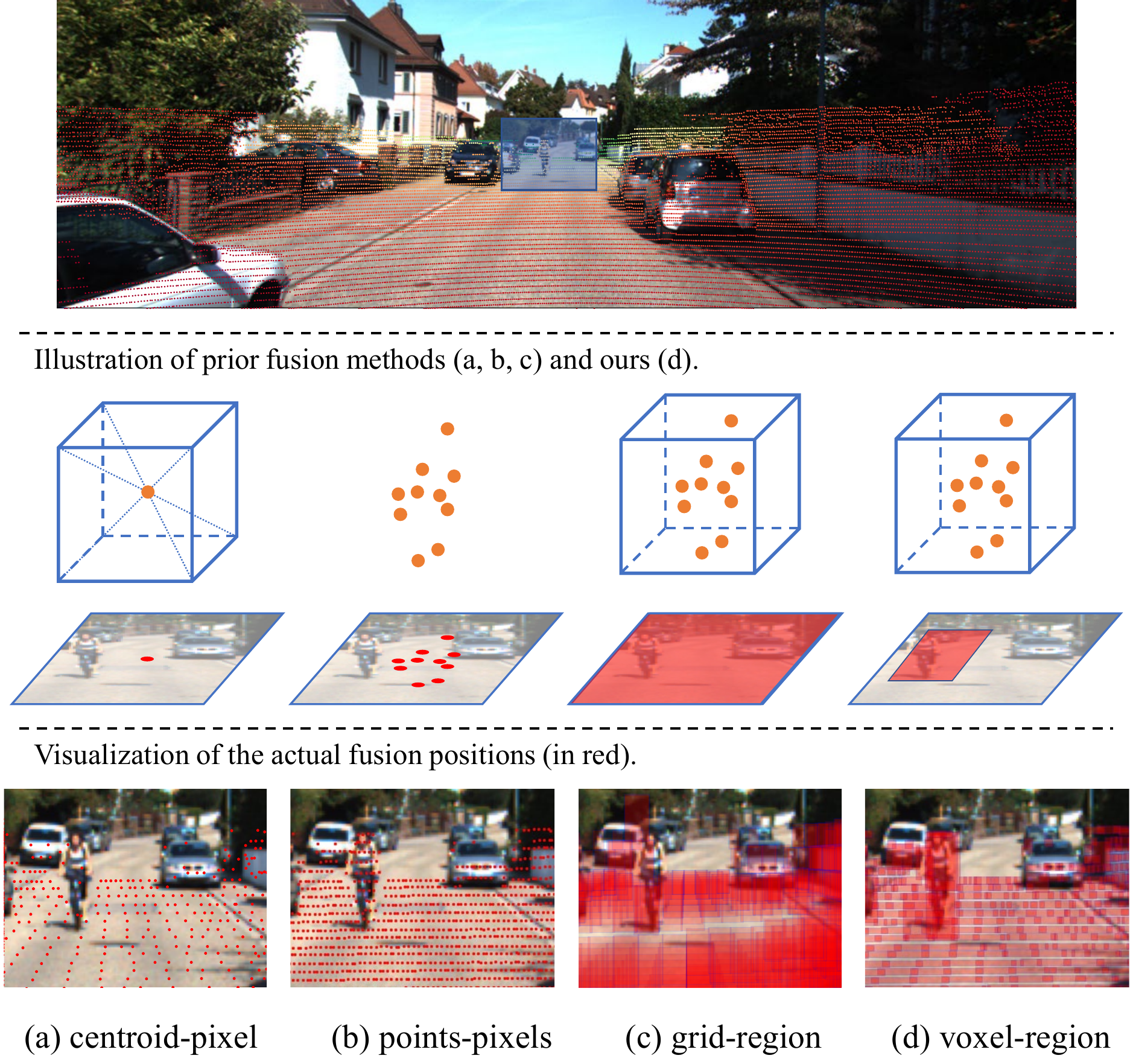}}
\caption{
\textbf{Top:} an image with point cloud projected. \textbf{Middle:} illustration of different fusion strategies, where the blue cubes are voxel grids, the orange points are point clouds, and the red areas are gathered pixels on the image plane. \textbf{Bottom:} visualization of the corresponding fusion strategies. \textbf{(a)}: Fusion by projecting the voxel centroid and interpolation, which only samples a few pixels and is prone to misalignment. \textbf{(b)}: Fusion in a point-to-pixel manner. The sampled pixels are still sparse, especially for small objects. \textbf{(c)}: Fusion by projecting the evenly spaced voxel grids causes misalignment between modalities and introduces too much background noise. The darker red color indicates more overlaps between regions. \textbf{(d)}: The voxel-region fusion we proposed to establish dense correspondence between a voxel and the image region while reducing background noise. The lighter red color indicates fewer overlaps and less noise.
}
\label{fig_compare}
\end{figure}

To deal with this data characteristic, researchers have proposed many fusion strategies. But these strategies still face the sparsity problem \cite{wu2022sparsefusedense}. Most of these methods can be classified according to the granularity of fusion \cite{wang2021multi}. As illustrated in Fig. \ref{fig_compare} (a), the voxel-level methods usually project the voxel centroids to the image plane to sample the image features and assign them to voxel features \cite{liang2018deepcontinuous,an2022deepstructural,li2022homogeneous,lin20223ddfm}. Here, we define them as centroid-pixel fusion. Due to ignoring the space that the voxels occupy, the sampled pixels are very sparse. Similarly, point-level methods typically associate one point with only a few image pixels \cite{sindagi2019mvx,huang2020epnet,vora2020pointpainting,xie2020pircnn,wang2021pointaugmenting,li2022deepfusion,wu2022sparsefusedense,zheng2022pifnet}, i.e., points-pixels fusion. As a result, the sampled pixels are equally sparse as the original point cloud, as seen in Fig. \ref{fig_compare} (b). Both of these two strategies are Sparse-to-Sparse fusion. MVX-Net \cite{sindagi2019mvx} proposed a denser fusion strategy, which projects the evenly spaced grid of voxels to the image plane and uses the RoI-pooling operation to sample image features, i.e., grid-region fusion. However, as illustrated in Fig. \ref{fig_compare} (c), grids with fixed sizes do not consider the distribution of points inside voxels. This results in more background image pixels associated with voxel features. Meanwhile, the projected regions have many overlaps with each other, causing a lot of confusion. As seen in Fig. \ref{fig_compare} (c), the darker red area indicates more overlaps. Another granularity is the RoI level. These methods \cite{chen2017mv3d,wang2019frustum,liang2019multitask,yoo20203dcvf} usually fuse RoI features based on 2D or 3D proposals in the late stage. Though the fusion granularity is denser, the performance of the whole model is likely to be limited by a single branch \cite{wang2021multi}.

Focusing on developing a dense and effective fusion strategy, we propose a novel Sparse-to-Dense Voxel Region Fusion (SDVRF) method to boost the representation ability of fused features with rich contextual information, as shown in Fig. \ref{fig_compare} (d). Concretely, the points are first divided evenly in space without downsampling operation \cite{zhou2020end}. Then, we use the local point cloud in every voxel to obtain a shape-variable rectangular area, named Voxel Region (VR), based on the calibration matrix of LiDAR and the camera. Notably, different from previous methods, which project the size-fixed voxels, the proposed VR is obtained flexibly according to the distribution of local point clouds. This achieves better alignment between local point clouds and image features. We also propose a multi-scale fusion framework to extract more contextual information and deal with objects of various sizes. We summarize our approach with two contributions:

\textbf{Voxel Region Fusion Module.} We proposed a new module to aggregate features from LiDAR point cloud and image by establishing dense correspondences between sparse points and dense pixels. Points in a voxel are considered a local point cloud, occupying a region in 3D space. We calculate an appropriate bounding box, i.e., the VR, for projected points on the image plane. The deep features of different modalities are fused within the region. Instead of attaching image features to points by sampling or projecting the size-fixed grids, we gather more spatial context for the local point cloud while avoiding introducing too much noise. We also use a pre-trained semantic segmentation backbone \cite{lraspp2019} to obtain image feature maps, reducing noise from images and accelerating convergence.

\textbf{Multi-scale Voxelization Fusion Framework.} Voxel-based methods usually face the problem of choice of voxel size. Previous LiDAR-based method voxelizes the point cloud by different sizes \cite{ye2020hvnet}. We further integrate local point clouds with image regions in various scales based on the proposed VRFM. Our network can generate the Bird's-Eye View (BEV) features with rich geometric and contextual information.

\section{Related Work}
\subsection{3D Object Detection Using LiDAR point clouds}

LiDAR-based 3D object detection aims to predict target objects' class, location, and direction from point clouds. LiDAR point clouds are irregular data with sparse distribution, and the number of points varies depending on the sample. Thus, unique feature extraction methods are needed to make them processable by deep-learning models. Most methods can be classified into three categories. Point-based methods \cite{yang2018pixor,yang2018ipod,shi2019pointrcnn,shi2020pvrcnn,shi2020pointgnn,yang20203dssd,zheng2021sessd,zhang2022notequal} adapt down-sampling to unify the number of points and ball query operation \cite{qi2017pointnet} to gather local features. These methods have a relatively flexible receptive field but may suffer information loss during sampling. Voxel-based methods \cite{lang2019pointpillars,deng2021voxelrcnn} first divide the point cloud into evenly sized grids and then extract features within each voxel. PointPillars \cite{lang2019pointpillars} only voxelizes on BEV, thus can reduce computation. Voxel R-CNN \cite{deng2021voxelrcnn} is proposed recently to refine detection based on coarse voxel features. These methods have higher computational efficiency, but the ability of local feature extraction is limited by pre-defined voxel size. Point-voxel-based methods try to combine their strengths. HVNet \cite{ye2020hvnet} is proposed to fuse multi-scale voxel feature and point feature at point-level. PV-RCNN combines 3D voxel Convolutional Neural Network (CNN) with PointNet-based set abstraction operations \cite{qi2017pointnet}. Other works \cite{guan2022m3detr} combine different point cloud representations. 

\subsection{LiDAR-camera fusion for 3D Object Detection}
There is a vast modal gap between LiDAR point cloud and image data. A point cloud consists of sparsely irregularly distributed 3D coordinate data, while an image is a compact combination of pixels. LiDAR-camera-based methods usually focus on researching better fusion strategies. Fusion can be performed at different granularity \cite{wang2021multi}. Most methods can be classified into RoI-, point-, and voxel-wise. Specifically, RoI-wise methods usually depend on a single modal branch because fusion happens between 2D or 3D proposals. Point-wise fusion is more commonly used. MVX-Net \cite{sindagi2019mvx} projects each point onto the image plane and appends the point with the image feature corresponding to the projected location index. PointPainting \cite{vora2020pointpainting} is proposed to attach image segmentation scores onto points before they are fed into the network. EPNet \cite{huang2020epnet} uses many Set Abstraction (SA) layers to extract key point features and fuse them with interpolated image features. These methods fuse in a point-to-pixel manner, which is a sparse correspondence. As for voxel-wise methods, MVX-Net \cite{sindagi2019mvx} projects equally spaced grids before voxelization. DVF \cite{mahmoud2023dense} samples image feature with projected voxel centers. These methods are either prone to introducing background noise or easily affected by alignment quality.

\section{PROPOSED METHOD}
\subsection{Framework Overview}

\begin{figure*}[htp]
\centerline{\includegraphics[width=0.9\textwidth]{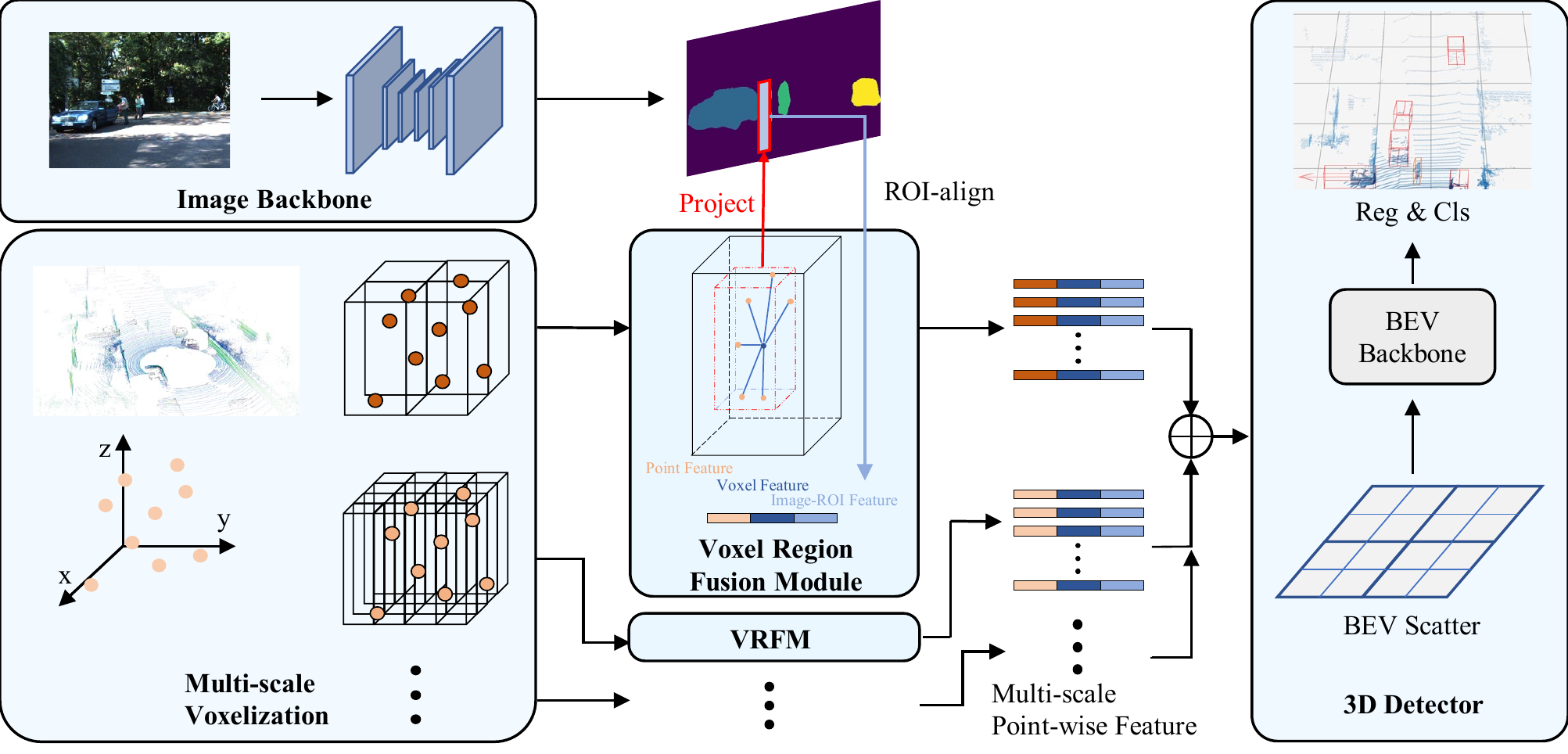}}
\caption{
Framework Overview. Each image is processed by a pre-trained semantic segmentation backbone to generate a feature map. The point cloud is voxelized with various scales and fed into the Voxel Region Fusion Modules (VRFMs, details in Section \ref{VRFM}) respectively. The local point cloud (in red dashed line) within each voxel is projected to obtain a region (in red rectangle) on the image plane dynamically, i.e., the Voxel Region. The multi-modal and multi-scale features are fused in the point level first and then converted to the voxel level (Section \ref{MSVF}). Finally, the boosted voxel features are fed into various 3D detectors to predict the final results.
}
\label{fig_overview}
\end{figure*}

The overall architecture of the proposed SDVRF is illustrated in Fig. \ref{fig_overview}. First, the LiDAR point cloud and the matched monocular camera image are processed by separate streams. Multi-scale voxelization operation is applied to all the points and assigns every point to predefined voxels of different scales. The image is fed into a pre-trained image segmentation backbone to generate a semantic feature map. Then we gather and fuse them using VRFM within different scales. The VRFM can generate multi-modal features by establishing dense and appropriate correspondences for every occupied non-empty voxel. Further, features from different modalities are concatenated and scattered to BEV to recover a dense BEV feature map, then processed by a commonly used 3D detector, usually composed of a BEV backbone and a detection head. In the following, we present the technical details of the proposed method.

\subsection{Image and LiDAR pipeline}
Pixels are densely arranged in an image, and their rich semantic information can be gathered by convolution operations. However, LiDAR point clouds are sparse, irregularly distributed, and have highly variable point densities. Before feeding them into the proposed VRFM, we apply different pipelines as illustrated in Fig. \ref{fig_overview}. 

For images captured by a monocular camera, we train a semantic segmentation backbone, i.e., LR-ASPP \cite{howard2019searching}, to generate image feature maps $\mathbf{F}_{img} \in \mathbb{R}^{H \times W \times C}$. Each channel in the image feature map represents the confidence score for a specific category, including background. Notice that the parameters of these layers are frozen during training. For LiDAR point clouds, voxelization can divide points into an evenly spaced grid of voxels. The size of a voxel is predefined as $(l, w, h)$, representing length along $X, Y, Z$ axes in 3D space. Following MVF \cite{zhou2020mvf}, we use dynamic voxelization, which establishes the bi-directional mapping between points and voxels without sampling operation. We define different voxel sizes for the Multi-scale Voxelization Fusion Framework and apply voxelization within every scale. Point $\mathbf{p}_i$ is assigned to voxels $\left\{\mathbf{v}_{j_0}, \ldots, \mathbf{v}_{j_s}\right\}$ based on their relative spatial coordinates in different scales.

\subsection{Voxel Region Fusion Module}
\label{VRFM}
Previous voxel-based fusion methods usually project the voxel centroids \cite{yoo20203dcvf, mahmoud2023dense} or the size-fixed grids \cite{sindagi2019mvx} to sample the image feature. The former leads to sparse points fused with sparse pixels, while the latter introduces too much background noise. We further propose the VRFM to achieve a balance between generating dense fusion and introducing noise.

\textbf{Projection from 3D to 2D.} The VRFM takes all the points of a point cloud in field of view $\mathbf{P}=\left\{\mathbf{p}_1, \ldots, \mathbf{p}_N\right\}$, their corresponding voxel indexes $\mathbf{I}=\left\{\mathbf{i}_1, \ldots, \mathbf{i}_N\right\}$ in a single scale, and the image feature map $\mathbf{F}_{I}$ as input. We use the transformation matrix $\mathbf{M}_{tran}$ to transform the points from the world coordinate to the camera coordinate. Then we use the camera's intrinsic matrix $\mathbf{M}_{intr}$ to project 3D points in the camera coordinate to the image plane. The projection process is expressed as
\begin{equation}
\mathbf{P}_{pix}=\mathbf{M}_{intr} \times \mathbf{M}_{tran} \times \mathbf{P} \div \mathbf{Z}_{c}
\end{equation}
where $\mathbf{P}_{pix}$ are homogeneous points in the pixel coordinates and $\mathbf{Z}_{c}$ is the depth of points in the camera coordinate. The $\div$ denotes element-wise division.

The above equations establish correspondences between the point cloud and the pixels. However, this projection heavily depends on the quality of the camera calibration matrices, including $\mathbf{M}_{intr}$ and $\mathbf{M}_{tran}$. Slight deviations can directly cause projection misalignment, causing 3D points to be assigned to the wrong pixels. Not only that, some fusion methods are also influenced by the local point cloud distribution within voxels. \cite{mahmoud2023dense, yoo20203dcvf} project the centroids of the size-fixed voxels, which is more likely to cause misalignment issues. \cite{sindagi2019mvx} compute eight corners of each voxel and get a 2D RoI on the image. The projection of evenly spaced grids is effective but introduces too much background noise.

\textbf{Generation of the Voxel Region.} Based on the projection process above, we present the generation of VR. A local point cloud usually occupies a sub-space of the whole voxel space, defined as a 3D region. After projection, we get the 2D points $\mathbf{P}_{pix}$ with one-to-one correspondence with the voxel indexes $\mathbf{I}$. The points within the same voxel share the same index. The number of points within each voxel and their location in local space varies. 

With the help of the scatter max and min operations, we dynamically generate rectangle regions on the image plane for each voxel. The 2D region is defined by two corners $(x_{min}, y_{min}, x_{max}, y_{max})$, i.e., the axis-aligned minimum bounding box of the 2D points set. All the projected points of the local point cloud are located in this 2D region. This can be formularized as follows:
\begin{equation}
\mathbf{R}_j = \underset{i}{Min} \left \{ \left ( x_i, y_i \right ) \mid I_i=j \right \} \oplus \underset{i}{Max} \left \{ \left ( x_i, y_i \right ) \mid I_i=j \right \} 
\end{equation}
where $\mathbf{R}_j$ is the generated 2D region of voxel $\mathbf{V}_j$, and $\oplus$ denotes concatenation operation.

To obtain more contextual information for voxels, we take the distance to the ego of 3D points into consideration. Points far away from the LiDAR are more sparse, which causes the number of points within a single voxel to be too small. This may lead to failure to generate the 2D region under special circumstances, e.g., only one point in a voxel. We use a scaling factor to enlarge the width and length to generate 2D regions with appropriate areas. The range of the point clouds is defined as $\left [ x_{min}, y_{min}, z_{min}, x_{max}, y_{max}, z_{max} \right ]$, and the ego is at the origin of the LiDAR coordinate system. We compute the scale factor for the voxel $V_j$ as
\begin{equation}
\alpha_j=1+\frac{\parallel \left ( x_j, y_j \right ) \parallel }{\parallel \left ( x_{max}, y_{max} \right ) \parallel} 
\end{equation}
where $\left ( x_j, y_j \right )$ is the centroid of the voxel $V_j$ in BEV. The width $w_j$ and length $l_j$ of the Voxel Region are enlarged by multiplying the scale factor as 
\begin{equation}
\begin{cases}
w_j=\alpha_j \left ( w_j + \delta \right ) \\
l_j=\alpha_j \left ( l_j + \delta \right )
\end{cases}
\end{equation}
where $\delta$ is the pre-defined offset to enlarge the region with zero area.

\begin{figure}[tp]
\centerline{\includegraphics[width=0.45\textwidth]{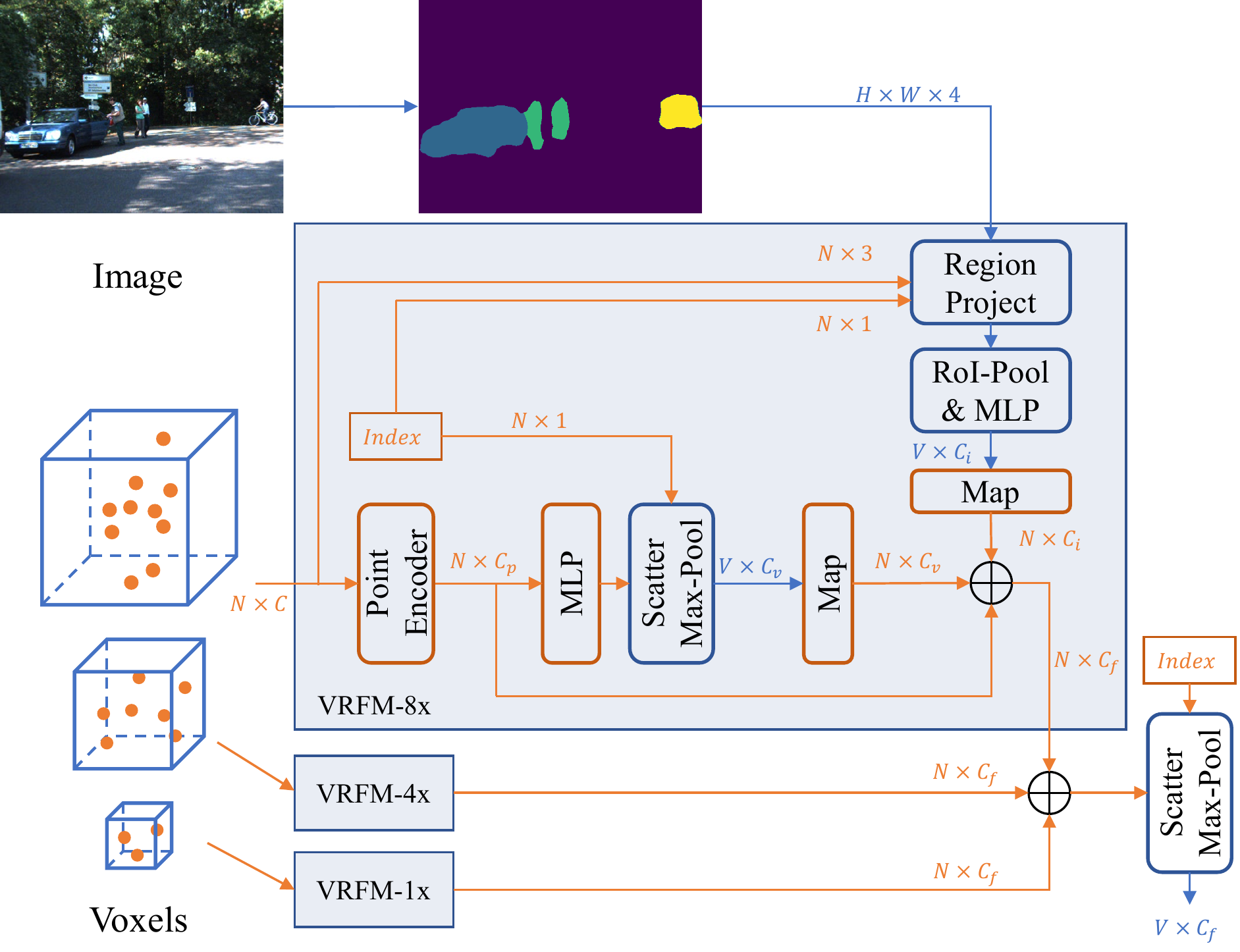}}
\caption{
Structure of the VRFM and the Multi-scale Voxelization Fusion Framework. Points within each voxel are encoded to voxel-level (\textcolor[RGB]{68,114,196}{blue}) features by a PointNet-based module, including MLP and max-pooling. The points' coordinates and their indexes of voxels are fed into the Region Project module to generate corresponding regions and pooled into voxel-level image features. Afterward, these features are mapped back to point-level (\textcolor[RGB]{237,125,49}{orange}) and fused by concatenation. At last,  the fusion features from various scales are concatenated and pooled to the voxel level.
}
\label{fig_net}
\end{figure}

\textbf{Extraction of the VR features.} After the generation of the 2D region $\left \{ {R}_j, j \in \mathbf{I} \right \}$, we extract features from different modalities, illustrated in Fig. \ref{fig_net}. Following \cite{lang2019pointpillars}, the point features are encoded as $\mathbf{F}_P = P \oplus P_c$, shaped $N \times C_p$, where $P_c$ is the relative coordinates with the centroids of voxels as the origin. The voxel features $\mathbf{F}_V \in \mathbb{R}^{V \times C_v}$ are extracted from point features by a simplified version of PointNet \cite{qi2017pointnet}, which includes a linear layer followed by Batch-Norm, ReLU, and a max-pooling operation. The image voxel features  $\mathbf{F}_I \in \mathbb{R}^{V \times C_i}$ are extracted from the image feature map using the voxel region. Specifically, the image voxel feature of the voxel $\mathbf{V}_j$ is sampled by the RoI-align operation to create an output tensor of fixed size, i.e., a tensor shaped $C_i \times 7 \times 7$. Then the feature is flattened and reduced along channels by a linear layer followed by Batch-Norm and ReLU. This process is formulated as
\begin{equation}
\mathbf{F}_I^j=MLP \left ( Flat \left ( Pool \left ( \mathbf{F}_{img}, \mathbf{R}_j\right ) \right ) \right )
\end{equation}

Then we fuse the features at point level by concatenation:
\begin{equation}
\mathbf{F}_{fuse}=\mathbf{F}_P \oplus M \left ( \mathbf{F}_V \right ) \oplus M \left ( \mathbf{F}_I \right )
\end{equation}
where $M$ denotes the mapping operation from voxels to points using indexes $\mathbf{I}$. 

The above algorithm we proposed differs from existing work. The points obtained by LiDAR scanning are discrete and sparse. General voxel-level fusion methods \cite{yoo20203dcvf,mahmoud2023dense}, which project the centroids of voxels (in Fig. \ref{fig_compare} (a)), face the problem of sparsity. Moreover, other methods \cite{sindagi2019mvx} project the equally spaced grids introduce too much background noise, especially when points in a voxel are few or gather in a corner (in Fig. \ref{fig_compare} (c), the red area is darker).

However, our method achieves a moderate fusion. By establishing a dense mapping between a local point cloud and a dense image region, VRFM can gather contextual information within the appropriate range while avoiding introducing too much background noise. Dynamically generated regions have little overlap with each other (in Fig. \ref{fig_compare} (d), the red area is lighter). Thus the problem of feature blurring \cite{vora2020pointpainting} is alleviated. Moreover, VRFM is less sensitive to projection misalignment. When points of targets are projected to the background on the image caused by deviation, general sampling strategies may extract incorrect features. Still, VRFM is able to obtain correct regional features based on contextual information.

\begin{table*}[htbp]
  \centering
  \caption{3D detection results on the KITTI test set using $\left.AP\right|_{R 40}$ metric for multi-class. The mAP-S refers to results on small objects. Best in \textcolor{blue}{blue}, and the second best in \textcolor{green}{green}.}
  \renewcommand\arraystretch{1.2}
  \resizebox{\linewidth}{!}{
    \begin{tabular}{c|l|c|cccc|cccc|cccc|c|c}
    \hline
    \multirow{2}{*}{Modality} & \multicolumn{1}{c|}{\multirow{2}{*}{Methods}} & \multirow{2}{*}{Reference} & \multicolumn{4}{c|}{Car(\%)}  & \multicolumn{4}{c|}{Pedestrian(\%)} & \multicolumn{4}{c|}{Cyclist(\%)} & \multirow{2}{*}{mAP-S(\%)} & \multirow{2}{*}{mAP(\%)} \\
\cline{4-15}          &       &       & Easy  & Mod.  & Hard  & mAP   & Easy  & Mod.  & Hard  & mAP   & Easy  & Mod.  & Hard  & mAP   &       &  \\
    \hline
    \multirow{6}[2]{*}{Single} & PointPillars \cite{lang2019pointpillars} & CVPR2019 & 82.58  & 74.31  & 68.99  & 75.29  & 51.45  & 41.92  & 38.89  & 44.09  & 77.10  & 58.65  & 51.92  & 62.56  & 53.32  & 60.65  \\
          & PointRCNN \cite{shi2019pointrcnn} & CVPR2020 & 85.94  & 75.76  & 68.32  & 76.67  & 49.43  & 41.78  & 38.63  & 43.28  & 73.93  & 59.60  & 53.59  & 62.37  & 52.83  & 60.78  \\
          & PV-RCNN \cite{shi2020pvrcnn} & CVPR2020 & 90.25  & 81.43  & 76.82  & 82.83  & \textcolor{green}{52.17}  & \textcolor{green}{43.29}  & \textcolor{green}{40.29}  & \textcolor{green}{45.25}  & 78.60  & 63.71  & 57.65  & 66.65  & 55.95  & \textcolor{green}{64.91}  \\
          & Voxel R-CNN \cite{deng2021voxelrcnn} & AAAI2021 & 90.90  & 81.62  & 77.06  & 83.19  & -     & -     & -     & -     & -     & -     & -     & -     & -     & - \\
          & FocalsConv \cite{chen2022focal} & CVPR2022 & 90.20  & 82.12  & 77.50  & 83.27  & -     & -     & -     & -     & -     & -     & -     & -     & -     & - \\
          & PDV \cite{hu2022pdv}   & CVPR2022 & 90.43  & 81.86  & 77.36  & 83.22  & 47.80  & 40.56  & 38.46  & 42.27  & \textcolor{blue}{83.04} & \textcolor{blue}{67.81}  & \textcolor{blue}{60.46}  & \textcolor{blue}{70.44}  & \textcolor{green}{56.36}  & \textcolor{blue}{65.31}  \\
    \hline
    \multirow{7}[2]{*}{Multi} & 3D-CVF \cite{yoo20203dcvf} & ECCV2020 & 89.20  & 80.05  & 73.11  & 80.79  & -     & -     & -     & -     & -     & -     & -     & -     & -     & - \\
          & PointPainting \cite{vora2020pointpainting} & CVPR2020 & 82.11  & 71.70  & 67.08  & 73.63  & 50.32  & 40.97  & 37.87  & 43.05  & 77.63  & 63.78  & 55.89  & 65.77  & 54.41  & 60.82  \\
          & Focals Conv-F \cite{chen2022focal} & CVPR2022 & 90.55  & 82.28  & \textcolor{green}{77.59}  & 83.47  & -     & -     & -     & -     & -     & -     & -     & -     & -     & - \\
          & SFD \cite{wu2022sparsefusedense}   & CVPR2022 & \textcolor{blue}{91.73}  & \textcolor{blue}{84.76}  & \textcolor{blue}{77.92}  & \textcolor{blue}{84.80}  & -     & -     & -     & -     & -     & -     & -     & -     & -     & - \\
          & EPNet++ \cite{liu2022epnet++} & TPAMI2022 & \textcolor{green}{91.37}  & 81.96  & 76.71  & 83.35  & \textcolor{blue}{52.79}  & \textcolor{blue}{44.38}  & \textcolor{blue}{41.29}  & \textcolor{blue}{46.15}  & 76.15  & 59.71  & 53.67  & 63.18  & 54.67  & 64.23  \\
          & DVF \cite{mahmoud2023dense}   & WACV2023 & 90.99  & \textcolor{green}{82.40}  & 77.37  & \textcolor{green}{83.59}  & -     & -     & -     & -     & -     & -     & -     & -     & -     & - \\
          & \textbf{Voxel R-CNN+ours} & -     & 88.48  & 79.48  & 74.73  & 80.90  & 50.22  & 42.79  & 39.37  & 44.13  & \textcolor{green}{82.56}  & \textcolor{green}{66.12}  & \textcolor{green}{59.55}  & \textcolor{green}{69.41}  & \textcolor{blue}{56.77}  & 64.81  \\
    \hline
    \end{tabular}%
    }
  \label{test_3D}%
\end{table*}%

\begin{table*}[t]
  \centering
  \caption{3D detection results on the KITTI val set using $\left.AP\right|_{R 40}$ metric for multi-class. The mAP-S refers to results on small objects. Best in \textcolor{blue}{blue}, and the second best in \textcolor{green}{green}. *denotes re-produced results.}
  \renewcommand\arraystretch{1.2}
  \resizebox{\linewidth}{!}{
    \begin{tabular}{c|l|c|cccc|cccc|cccc|c|c}
    \hline
    \multirow{2}{*}{Modality} & \multicolumn{1}{c|}{\multirow{2}{*}{Methods}} & \multirow{2}{*}{Reference} & \multicolumn{4}{c|}{Car(\%)}  & \multicolumn{4}{c|}{Pedestrian(\%)} & \multicolumn{4}{c|}{Cyclist(\%)} & \multicolumn{1}{c|}{\multirow{2}{*}{mAP-S(\%)}} & \multirow{2}{*}{mAP(\%)}\\
\cline{4-15}          &       &       & Easy  & Mod.  & Hard  & mAP   & Easy  & Mod.  & Hard  & mAP   & Easy  & Mod.  & Hard  & mAP   &       &  \\
    \hline
    \multirow{6}{*}{Single} & PointPillars* \cite{lang2019pointpillars} & CVPR2019 & 86.83  & 78.09  & 75.44  & 80.12  & 57.44  & 51.95  & 47.14  & 52.18  & 80.28  & 63.07  & 58.71  & 67.35  & 59.77  & 66.55  \\
          & PointRCNN* \cite{shi2019pointrcnn} & CVPR2020 & 91.67  & 80.56  & 78.08  & 83.44  & 62.66  & 54.70  & 48.07  & 55.14  & 91.77  & 72.79  & 68.36  & \textcolor{green}{77.64}  & 66.39  & 72.07  \\
          & PV-RCNN* \cite{shi2020pvrcnn} & CVPR2020 & 92.10  & 84.36  & 82.48  & 86.31  & 64.26  & 56.67  & 51.91  & 57.61  & 88.88  & 71.95  & 66.78  & 75.87  & 66.74  & 73.27  \\
          & Voxel R-CNN* \cite{deng2021voxelrcnn} & AAAI2021 & 89.68  & 81.05  & 78.84  & 83.19  & 68.11  & 62.26  & 57.09  & 62.49  & 90.17  & 72.14  & 67.76  & 76.69  & 69.59 
 & 74.12  \\
          & FocalsConv \cite{chen2022focal} & CVPR2022 & 92.32  & 85.19  & 82.62  & 86.71  & -     & -     & -     & -     & -     & -     & -     & -     & -     & - \\
          & PDV \cite{hu2022pdv}  & CVPR2022 & \textcolor{blue}{92.56}  & \textcolor{green}{85.29}  & \textcolor{blue}{83.05}  & \textcolor{blue}{86.97}  & 66.90  & 60.80  & 55.85  & 61.18  & \textcolor{green}{92.72}  & \textcolor{blue}{74.23}  & \textcolor{blue}{69.60}  & \textcolor{blue}{78.85}  & 70.02  & \textcolor{blue}{75.67}  \\
    \hline
    \multirow{7}{*}{Multi} & 3D-CVF \cite{yoo20203dcvf} & ECCV2020 & 89.67  & 79.88  & 78.47  & 82.67  & -     & -     & -     & -     & -     & -     & -     & -     & -     & - \\
          & PointPainting \cite{vora2020pointpainting} & CVPR2020 & 88.38  & 77.74  & 76.76  & 80.96  & 69.38  & 61.67  & 54.58  & 61.88  & 85.21  & 71.62  & 66.98  & 74.60  & 68.24  & 72.48  \\
          & Focals Conv-F \cite{chen2022focal} & CVPR2022 & 92.26  & \textcolor{blue}{85.32}  & 82.95  & 86.84  & -     & -     & -     & -     & -     & -     & -     & -     & -     & - \\
          & EPNet++ \cite{liu2022epnet++} & TPAMI2022 & \textcolor{green}{92.51}  & 83.17  & 82.27  & 85.98  & \textcolor{blue}{73.77}  & \textcolor{green}{65.42}  & \textcolor{green}{59.13}  & \textcolor{blue}{66.11}  & 86.23  & 63.82  & 60.02  & 70.02  & 68.07  & 74.04  \\
          & DVF \cite{mahmoud2023dense}  & WACV2023 & 92.34  & 85.25  & \textcolor{green}{82.97}  & \textcolor{green}{86.85}  & 66.08  & 59.18  & 54.68  & 59.98  & 90.93  & 72.46  & 68.05  & 77.15  & 68.56  & 74.66  \\
          & \textbf{PointPillars+ours} & -     & 88.95  & 77.85  & 75.22  & 80.67  & 68.18  & 62.75  & 57.79  & 62.91  & \textcolor{blue}{93.27}  & 71.60  & 67.34  & 77.41  & \textcolor{green}{70.16}  & 73.66  \\
          & \textbf{Voxel R-CNN+ours} & -     & 88.89  & 80.71  & 78.47  & 82.69  & \textcolor{green}{71.07}  & \textcolor{blue}{65.53}  & \textcolor{blue}{60.16}  & \textcolor{green}{65.59}  & 89.57  & \textcolor{green}{73.09}  & \textcolor{green}{68.59}  & 77.08  & \textcolor{blue}{71.34}  & \textcolor{green}{75.12}  \\
    \hline
    \end{tabular}%
    }
  \label{val_3D}%
\end{table*}%

\subsection{Multi-scale Voxelization Fusion Framework}
\label{MSVF}
The PointNet-based encoder in most voxel-based methods operates on the unordered local point cloud and computes a unified feature vector by max-pooling. Thus, its ability to aggregate the local geometric and semantic information is associated with the space size of the voxel. Motivated
by recent works \cite{shi2020pvrcnn, mahmoud2023dense}, which obtain multi-scale features by 3D convolution, we further proposed a Multi-scale Voxelization Fusion Framework based on the VRFM.

Voxelization divides the point clouds into evenly spaced grids, and the number of voxels is determined directly by the size of each voxel. Following HVNet \cite{ye2020hvnet}, which encodes voxel information from different scales into the point-wise feature, we fuse multi-modal features in various voxel scales with the help of VRFM. 

As mentioned above, the voxel size is defined as $(l, w, h)$. We further define a set of voxel scales $S=\left \{ s \mid s \in \left [ 1, N_s \right ] \right \}$. The voxel size in scale $s$ is $(l*s, w*s, h*s)$. 

When $s=1$, smaller voxels lead to a higher resolution of the BEV feature map, which can keep more detailed characteristics of small objects. However it is hard to obtain enough local geometric information. Fewer points in a voxel, e.g., less than five, lead to a simple local spatial structure, suppressing the representation ability of the voxel feature. Moreover, the area of the generated VR is also tiny, causing many foreground pixels not to be gathered.

When $s>1$, the voxels are enlarged to gather better local features, especially geometric information, benefiting from larger receptive fields. 
The proposed VRFM projects larger 3D space occupied by the local point cloud to get VRs, which are enlarged, too. The distribution of point clouds is sparse. With a larger voxel size, pixels in the gap between points are captured by a larger 2D region on the image plane, i.e., the image feature extracted by VRFM for each voxel includes more contextual information.

The fusion features in the scale $s$ extracted by VRFM are defined as $\mathbf{F}_s \in \mathbb{R}^{N \times C}$. We concatenate fusion features in all scales point-wise as:
\begin{equation}
    \mathbf{F}_\mathbf{S}=concat\left \{ \mathbf{F}_s \mid s \in \left [ 1, N_s \right ] \right \}
\end{equation}
where the fusion feature $\mathbf{F}_\mathbf{S}$ incorporates the basic point feature, the voxel feature encoded by the PointNet-based encoder, and the image region feature extracted from the projected 2D regions in various scales. To detect objects in BEV, we use a PointNet layer to convert the sparse point-wise fusion features to voxel-level features $\mathbf{F}_\mathbf{SV} \in \mathbb{R}^{V \times C}$. Then we scatter them to BEV space and use prevalent BEV backbone and 3D object detection heads for the final prediction. The proposed Multi-scale Voxelization Fusion Framework can be applied to any voxel-based LiDAR backbone.
\begin{table}[tp]
  \centering
  \caption{Improvements on the KITTI val set using $\left.AP\right|_{R 40}$ metric for Pedestrian and Cyclist. *denotes re-produced results.}
  \renewcommand\arraystretch{1.2}
  \resizebox{\linewidth}{!}{
    \begin{tabular}{l|cc|c|cc|c}
    \hline
    \multicolumn{1}{c|}{\multirow{2}{*}{Methods}} & \multicolumn{3}{c|}{3D(\%)} & \multicolumn{3}{c}{BEV(\%)} \\
\cline{2-7}          & Pedestrian & Cyclist & mAP   & Pedestrian & Cyclist & mAP \\
    \hline
    PointPillars* & 52.18  & 67.35  & 59.77  & 58.03  & 71.66  & 64.85  \\
    PointPillars+ours & 62.91  & 77.41  & 70.16  & 70.06  & 79.93  & 75.00  \\
    \textit{Improvement} & \textit{+10.73} & \textit{+10.05} & \textit{+10.39} & \textit{+12.02} & \textit{+8.28} & \textit{+10.15} \\
    \hline
    Voxel R-CNN* & 62.49  & 76.69  & 69.59  & 66.82  & 78.25  & 72.54  \\
    Voxel R-CNN+ours & 65.59  & 77.08  & 71.34  & 69.12  & 79.75  & 74.44  \\
    \textit{Improvement} & \textit{+3.10} & \textit{+0.39} & \textit{+1.75} & \textit{+2.30} & \textit{+1.50} & \textit{+1.90} \\
    \hline
    \end{tabular}%
    }
  \label{val_improve}%
\end{table}%

\begin{table*}[ht]
  \centering
  \caption{BEV detection results on the KITTI val set using $\left.AP\right|_{R 40}$ metric for multi-class. *denotes re-produced results.}
  \renewcommand\arraystretch{1.2}
  \resizebox{\linewidth}{!}{
    \begin{tabular}{l|cccc|cccc|cccc|c}
    \hline
    \multicolumn{1}{c|}{\multirow{2}{*}{Methods}} & \multicolumn{4}{c|}{Car(\%)}  & \multicolumn{4}{c|}{Pedestrian(\%)} & \multicolumn{4}{c|}{Cyclist(\%)} & \multirow{2}{*}{mAP(\%)} \\
\cline{2-13}          & Easy  & Mod.  & Hard  & mAP   & Easy  & Mod.  & Hard  & mAP   & Easy  & Mod.  & Hard  & mAP   &  \\
    \hline
    PV-RCNN* & 93.02  & 90.30  & 88.52  & 90.61  & 66.99  & 58.85  & 54.57  & 60.14  & 93.45  & 73.50  & 70.15  & 79.03  & 76.59  \\
    \hline
    PointPillars* & 90.86  & 87.31  & 86.28  & 88.15  & 63.06  & 57.57  & 53.47  & 58.03  & 84.51  & 67.43  & 63.03  & 71.66  & 72.61  \\
    PointPillars+ours & 93.15  & 86.89  & 84.63  & 88.22  & 75.26  & 69.69  & 65.22  & 70.06  & 95.02  & 74.56  & 70.22  & 79.93  & 79.41  \\
    \textit{Improvement} & \textit{+2.29} & \textit{-0.42} & \textit{-1.65} & \textit{+0.07} & \textit{+12.20} & \textit{+12.12} & \textit{+11.76} & \textit{+12.02} & \textit{+10.52} & \textit{+7.13} & \textit{+7.19} & \textit{+8.28} & \textit{+6.79} \\
    \hline
    Voxel R-CNN* & 90.89  & 87.05  & 86.91  & 88.29  & 72.37  & 66.52  & 61.57  & 66.82  & 91.23  & 73.28  & 70.25  & 78.25  & 77.79  \\
    Voxel R-CNN+ours & 91.54  & 86.66  & 86.49  & 88.23  & 74.53  & 68.95  & 63.89  & 69.12  & 92.76  & 74.88  & 71.61  & 79.75  & 79.03  \\
    \textit{Improvement} & \textit{+0.65} & \textit{-0.39} & \textit{-0.42} & \textit{-0.05} & \textit{+2.15} & \textit{+2.43} & \textit{+2.32} & \textit{+2.30} & \textit{+1.53} & \textit{+1.61} & \textit{+1.36} & \textit{+1.50} & \textit{+1.25} \\
    \hline
    \end{tabular}%
    }
  \label{bev}%
\end{table*}%

\begin{table}[ht]
  \centering
  \caption{Effects of different components on the KITTI val set using 3D $\left.AP\right|_{R 40}$ metric. Best in bold.}
  \renewcommand\arraystretch{1.2}
  \resizebox{\linewidth}{!}{
    \begin{tabular}{l|ll|ccc|c}
    \hline
    \multicolumn{1}{c|}{Baseline} & \multicolumn{1}{c}{VRFM} & \multicolumn{1}{c|}{Multi-scale} & Car(\%) & Pedestrian(\%) & Cyclist(\%) & mAP(\%) \\
    \hline
    \multirow{4}{*}{PointPillars} &       &       & 80.12  & 52.18  & 67.35  & 66.55  \\
          & \makecell[c]{\checkmark} &       & \textbf{81.34}  & 60.65  & 76.93  & 72.97 \\
          &       & \makecell[c]{\checkmark} & 79.68  & 60.42  & 74.83  & 71.64  \\
          & \makecell[c]{\checkmark} & \makecell[c]{\checkmark} & 80.67  & \textbf{62.91}  & \textbf{77.41}  & \textbf{73.66}  \\
    \hline
    \multirow{4}{*}{Voxel R-CNN} &       &       & 83.19  & 62.49  & 76.69  & 74.12  \\
          & \makecell[c]{\checkmark} &       & 82.91  & 64.41  & 75.79  & 74.37 \\
          &       & \makecell[c]{\checkmark} & \textbf{83.30}  & 62.42  & 76.96  & 74.23  \\
          & \makecell[c]{\checkmark} & \makecell[c]{\checkmark} & 82.69  & \textbf{65.59}  & \textbf{77.08}  & \textbf{75.12}  \\
    \hline
    \end{tabular}%
    }
  \label{ablation}%
\end{table}%

\begin{table}[ht]
  \centering
  \caption{Comparison between SDVRF and sparse-to-sparse fusion method on the KITTI val set using 3D $\left.AP\right|_{R 40}$ metric. Best in bold.}
  \renewcommand\arraystretch{1.2}
  \resizebox{\linewidth}{!}{
    \begin{tabular}{l|ll|ccc|c}
    \hline
    \multicolumn{1}{c|}{Baseline} & \multicolumn{1}{c}{Sparse-to-Sparse} & \multicolumn{1}{c|}{SDVRF} & Car(\%) & Pedestrian(\%) & Cyclist(\%) & mAP(\%) \\
    \hline
    \multicolumn{1}{c|}{\multirow{2}[2]{*}{PointPillars}} & \makecell[c]{\checkmark} &       & 79.87  & \textbf{63.25}  & 76.50  & 73.21  \\
          &       & \makecell[c]{\checkmark} & \textbf{80.67}  & 62.91  & \textbf{77.41}  & \textbf{73.66}  \\
    \hline
    \multirow{2}[2]{*}{Voxel R-CNN} & \makecell[c]{\checkmark} &       & \textbf{83.26}  & 63.72  & 76.66  & 74.55  \\
          &       & \makecell[c]{\checkmark} & 82.69  & \textbf{65.59}  & \textbf{77.08}  & \textbf{75.12}  \\
    \hline
    \end{tabular}%
    }
  \label{paint}%
\end{table}%

\section{EXPERIMENTS}
\subsection{Dataset and Evaluation Metrics}
We evaluate our method on the KITTI 3D and BEV object detection benchmark \cite{geiger2012kitti}, which includes 7,481 training samples and 7,518 testing samples. Following \cite{chen2015split}, the training samples are split into 3,712 and 3,769 samples for train and val set, respectively. The samples are stratified into easy, moderate, and hard difficulties according to bounding box height, occlusion, and truncation. We calculate the mAP on the val set using 40 recall positions $\left.AP\right|_{R 40}$. The IoU thresholds are set to 0.7, 0.5, and 0.5 for Car, Pedestrian, and Cyclist.

We use segmentation annotations \cite{heylen2021monocinis} for all 7481 training images to train the image backbone. The annotations include manually annotated vehicle and Pedestrian instances. We convert them to instance-agnostic annotations for semantic segmentation and use 2D bounding boxes to generate masks for Cyclists. The training samples are split for train and val set consistent with the KITTI dataset.

\subsection{Implementation Details}

 \textbf{Model setup.} We implement our network based on two LiDAR-only baselines, PointPillars \cite{lang2019pointpillars} and Voxel R-CNN \cite{deng2021voxelrcnn}. We reimplement them by the open-sourced OpenPCDet \cite{openpcdet2020} with dynamic voxelization operation \cite{zhou2020mvf} for a fair comparison. We follow their original perception range settings for different baselines: for PointPillars, the range of point cloud is [(0, 69.12), (-39.68, 39.68), (-3, 1)] meters and the base voxel size is [0.08, 0.08, 4] meters along the (x, y, z) axis; for Voxel R-CNN, we set them to [(0, 70.4), (-40, 40), (-3, 1)], and [0.05, 0.05, 0.1], respectively. We set the voxel scales for multi-scale experiments as $S=\left \{ 1, 4, 8 \right \}$. We follow the approaches mentioned in \cite{lang2019pointpillars} for LiDAR point clouds data augmentation and MoCa \cite{zhang2020moca} for cross-modal data augmentation. We use a lightweight network, LR-ASPP \cite{lraspp2019}, as the image semantic segmentation backbone, and the output channel is set to 4.
 
 \textbf{Training and inference.} We train the network on 2 GTX 3080 GPUs. The optimization strategy is consistent with PointPillars \cite{lang2019pointpillars} and Voxel R-CNN \cite{deng2021voxelrcnn}. During training, the LR-ASPP image backbone is frozen. We train a single model for detecting multi-class objects together and evaluate it without ensembling or testing augmentation.

\begin{figure*}[htp]
\centerline{\includegraphics[width=0.9\textwidth]{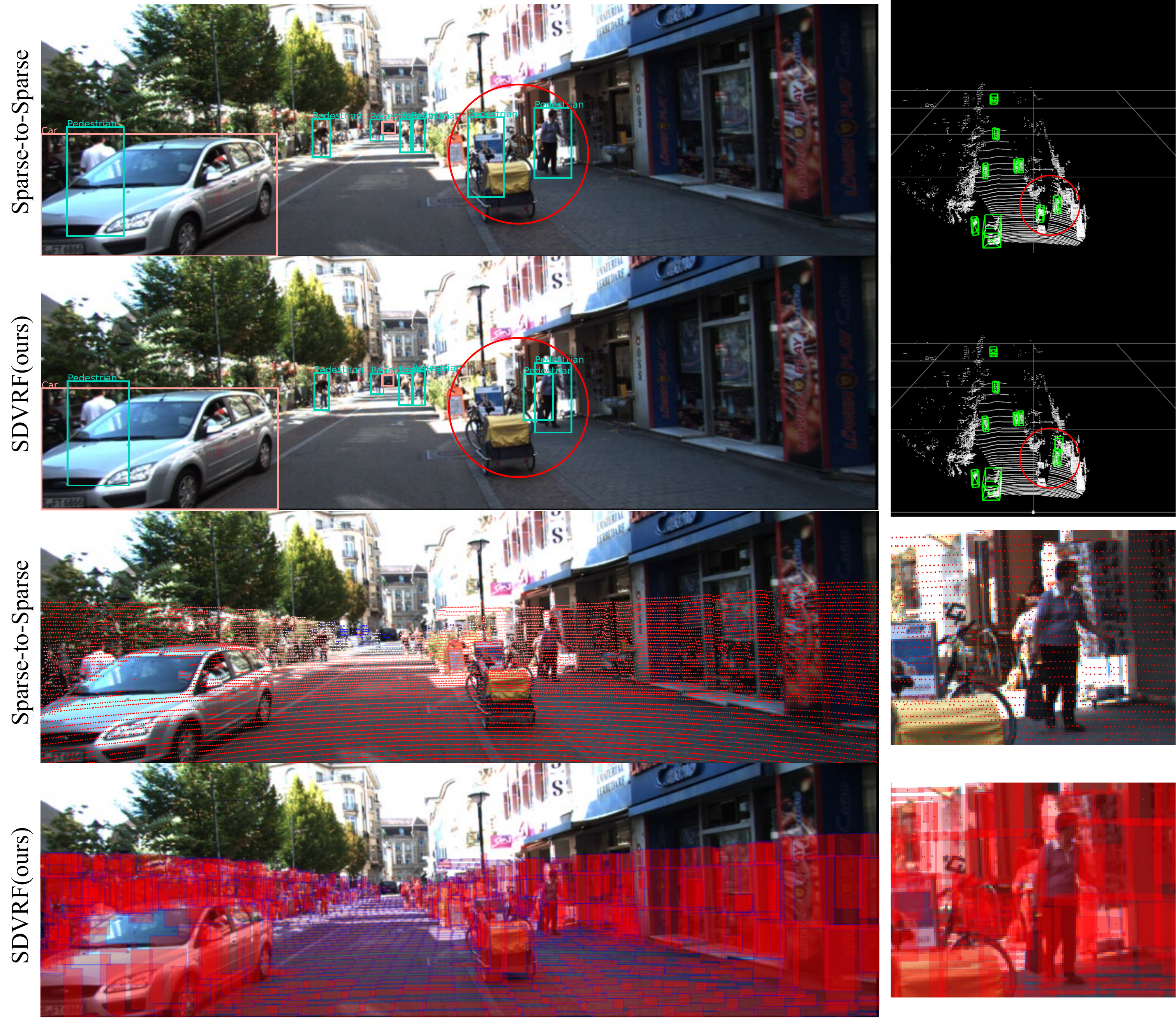}}
\caption{
Qualitative results on KITTI val set. The first two rows show a comparison of detection results between the Sparse-to-Sparse fusion and the proposed SDVRF. The Sparse-to-Sparse fusion method fails to detect an occluded pedestrian but outputs a False Positive. In comparison, the SDVRF is able to detect successfully (highlighted in red). The last two rows correspondingly display these two methods' actual fusion location or region (in red). The SDVRF is able to fuse within dense regions, thus extracting more pixels in the image. Best viewed with color and zoom-in.
}
\label{fig_visual}
\end{figure*}

\subsection{Comparison with State-of-the-Arts}

\textbf{3D Results on the KITTI dataset.} Table \ref{test_3D} presents the quantitative comparison with state-of-the-art LiDAR-only and multi-modal 3D object detection methods on the KITTI test set. The huge gap between representations of images and point clouds makes it difficult to fuse them, resulting in suboptimal performance. However, our method achieves Sparse-to-Dense fusion within the Voxel Region and gets better or comparable performance over single- and multi-modal methods, especially for small targets, i.e., Pedestrian and Cyclist. Notably, our method achieves 56.77\% mAP-S, outperforming the state-of-the-art multi-modal method EPNet++ \cite{liu2022epnet++} by 2.1\%. Our method achieves 64.81\% mAP on all classes, which is also comparable with LiDAR-only methods. Table \ref{val_3D} shows similar results on the val set. Our method achieves 71.34\% mAP-S, outperforming the state-of-the-art multi-modal method DVF \cite{mahmoud2023dense} by 2.78\%. Many samples of Pedestrian and Cyclist are particularly small or truncated. Thus the LiDAR points are extremely sparse, and a small degree of misalignment can cause much noise. The experiment results reveal that our method can effectively utilize the contextual information from images to locate these small objects while reducing noise, benefiting from the dynamic voxel region generation. Moreover, as shown in Table \ref{val_improve}, our method approach significantly enhances PointPillars \cite{lang2019pointpillars} and Voxel R-CNN \cite{deng2021voxelrcnn} baseline for small objects. Our method shows consistent improvement for both single-stage and two-stage single-modal detection baselines, demonstrating the effectiveness of multi-modal fusion.

\textbf{BEV Results on the KITTI val set.} BEV representation is suitable for the perception of the surrounding environment of cars benefited by near lack of occlusion. As shown in Table \ref{bev}, our method improves the accuracy of small objects in BEV for a large margin. As shown in Table \ref{val_improve}, our method significantly boosts the BEV mAP for both Pedestrian and Cyclist. The result on the single-stage baseline, i.e., PointPillars, achieves 75.00\% mAP, surpassing the result on the two-stage Voxel R-CNN, which is 74.44\%, by 0.56\%. The PointPillars does not divide space vertically but scatters the pillar features directly into BEV space without height compression. Compared with smaller voxels used in Voxel R-CNN, the local point cloud within each pillar contains more points and thus can include more complete geometrical characteristics. Meanwhile, with the help of the proposed VRFM, each pillar can be associated with a more suitable voxel region dynamically.

\textbf{Qualitative Results.} Based on the Voxel R-CNN baseline, we compare the detection results of the Sparse-to-Sparse fusion method and the proposed SDVRF in Fig. \ref{fig_visual}. The Sparse-to-Sparse fusion method samples pixels in a one-to-one manner \cite{vora2020pointpainting}, thus causing many pixels to be wasted on the image plane. As seen in the third row of \ref{fig_visual}, only pixels in red are sampled and fused with the point cloud, which does not change the sparsity of the LiDAR features. The SDVRF is able to leverage more pixels within the proposed VR to reduce the False Negative and False Positive, achieving Sparse-to-Dense fusion. As seen in the last row of \ref{fig_visual}, the pixels of image features in the red region are pooled and associated with the voxel features, and the pedestrian is fully covered. Some voxels that do not contain the target object correspond to very small regions on the image plane, such as the narrow areas on the ground. This demonstrates better alignment between local point clouds and image features.

\subsection{Ablation Study}
We perform ablation studies to verify the effect of the proposed VRFM and the multi-scale fusion framework on the KITTI val set. 

\textbf{Effect of each component.} The baselines use hard voxelization to process the point clouds. As shown in Table \ref{ablation}, we reimplement the two baselines and conduct all the experiments with dynamic voxelization \cite{zhou2020mvf}. The multi-scale component for the LiDAR-only network is reimplemented following HVNet \cite{ye2020hvnet}, and the mean voxel feature encoder of Voxel R-CNN is replaced. For PointPillars, which has a simple voxel feature encoder, the multi-scale component or the VRFM with single-scale both improve the performance for a large margin. This may be because the extracted information enriches the feature of a single pillar. Still, combing with the proposed VRFMs, the multi-scale fusion framework brings performance gains of 2.49\% and 2.58\% mAP over the single-modal results for Pedestrian and Cyclist, respectively. Image features from nearby areas at multiple scales are fused with voxel features generated by multi-scale voxelization operation. As for the Voxel R-CNN baseline, the VRFM component with single-scale reduces performance on Car and Cyclist. However, the multi-scale fusion framework with multi-modal input still improves the results on small objects and the multi-class mAP. It achieves a relative gain of 3.10\% and 0.39\% mAP for Pedestrian and Cyclist over the baseline, respectively. This strongly demonstrates the effectiveness of the proposed Multi-scale Voxelization Fusion Framework. With appropriate multi-modal and multi-scale contextual information, the network is able to locate small objects effectively.

\textbf{Comparison with Sparse-to-Sparse fusion.} The common fusion strategy of establishing one-to-one mappings between LiDAR points and image pixels is too sparse for appropriate fusion. A paradigm of Sparse-to-Sparse fusion is the element-wise concatenation of point features and the corresponding pixel features. In Table \ref{paint}, we compare this fusion method and our Sparse-to-Dense fusion method. The Sparse-to-Sparse networks are modified from the proposed SDVRF for a fair comparison. This fusion operation is implemented following \cite{vora2020pointpainting} by sampling pixels for every projected point to replace the VRFM. The other parts of the network and the training schedule remain the same. As shown in Table \ref{paint}, the mAP of the proposed network with VRFM outperforms the Sparse-to-Sparse fusion method on both two baselines. For small-size classes on the Voxel R-CNN baseline, our method surpasses the Sparse-to-Sparse method by a large margin. This strongly demonstrates that the dense correspondence and contextual information obtained by our fusion strategy can relieve the sparsity problem caused by small size and far distance.

\section{Conclusion and Limitation}
In this paper, we propose the Region Voxel Fusion Module to achieve a dense and effective fusion between sparse LiDAR points and dense image pixels. The contextual information from both modalities is integrated to enrich the voxel features in a Sparse-to-Dense manner, and the background noise is reduced, benefiting from better alignment. Moreover, we propose a Multi-scale Voxelization Fusion Framework to fuse local point clouds with image regions in various scales and deal with the problem of varying scales of target objects. Extensive experiments on the KITTI dataset show that the proposed SDVRF is able to achieve better performance on multi-modal 3D and BEV detection, especially for small objects like Pedestrian and Cyclist. Our method is able to be applied to other LiDAR-based methods, and we hope our method can serve as a better alternative to the current fusion strategies for multi-modal 3D objection. The limitation of our work is that we have not achieved high performance on cars that match the results on small objects. We will explore this to make the model more general in future works.

\section*{Acknowledgment}
This work was supported partly by the National Natural Science Foundation of China (Grant No. 62173045), partly by the Natural Science Foundation of Hainan Province (Grant No. 622RC675).
\bibliographystyle{unsrt}
\bibliography{ref}

\begin{thebibliography}{10}

\bibitem{mao20223d}
Jiageng Mao, Shaoshuai Shi, Xiaogang Wang, and Hongsheng Li.
\newblock 3d object detection for autonomous driving: a review and new
  outlooks.
\newblock {\em arXiv preprint arXiv:2206.09474}, 2022.

\bibitem{qian20223d}
Rui Qian, Xin Lai, and Xirong Li.
\newblock 3d object detection for autonomous driving: a survey.
\newblock {\em Pattern Recognition}, 130:108796, 2022.

\bibitem{wang2021multi}
Yingjie Wang, Qiuyu Mao, Hanqi Zhu, Jiajun Deng, Yu~Zhang, Jianmin Ji, Houqiang
  Li, and Yanyong Zhang.
\newblock Multi-modal 3d object detection in autonomous driving: a survey.
\newblock {\em arXiv preprint arXiv:2106.12735}, 2021.

\bibitem{wu2022sparsefusedense}
Xiaopei Wu, Liang Peng, Honghui Yang, Liang Xie, Chenxi Huang, Chengqi Deng,
  Haifeng Liu, and Deng Cai.
\newblock Sparse fuse dense: Towards high quality 3d detection with depth
  completion.
\newblock In {\em Proceedings of the IEEE/CVF Conference on Computer Vision and
  Pattern Recognition}, pages 5418--5427, 2022.

\bibitem{liang2018deepcontinuous}
Ming Liang, Bin Yang, Shenlong Wang, and Raquel Urtasun.
\newblock Deep continuous fusion for multi-sensor 3d object detection.
\newblock In {\em Proceedings of the European conference on computer vision
  (ECCV)}, pages 641--656, 2018.

\bibitem{an2022deepstructural}
Pei An, Junxiong Liang, Kun Yu, Bin Fang, and Jie Ma.
\newblock Deep structural information fusion for 3d object detection on
  lidar--camera system.
\newblock {\em Computer Vision and Image Understanding}, 214:103295, 2022.

\bibitem{li2022homogeneous}
Xin Li, Botian Shi, Yuenan Hou, Xingjiao Wu, Tianlong Ma, Yikang Li, and Liang
  He.
\newblock Homogeneous multi-modal feature fusion and interaction for 3d object
  detection.
\newblock In {\em Computer Vision--ECCV 2022: 17th European Conference, Tel
  Aviv, Israel, October 23--27, 2022, Proceedings, Part XXXVIII}, pages
  691--707. Springer, 2022.

\bibitem{lin20223ddfm}
Chunmian Lin, Daxin Tian, Xuting Duan, Jianshan Zhou, Dezong Zhao, and Dongpu
  Cao.
\newblock 3d-dfm: anchor-free multimodal 3-d object detection with dynamic
  fusion module for autonomous driving.
\newblock {\em IEEE Transactions on Neural Networks and Learning Systems},
  2022.

\bibitem{sindagi2019mvx}
Vishwanath~A Sindagi, Yin Zhou, and Oncel Tuzel.
\newblock Mvx-net: Multimodal voxelnet for 3d object detection.
\newblock In {\em 2019 International Conference on Robotics and Automation
  (ICRA)}, pages 7276--7282. IEEE, 2019.

\bibitem{huang2020epnet}
Tengteng Huang, Zhe Liu, Xiwu Chen, and Xiang Bai.
\newblock Epnet: Enhancing point features with image semantics for 3d object
  detection.
\newblock In {\em Computer Vision--ECCV 2020: 16th European Conference,
  Glasgow, UK, August 23--28, 2020, Proceedings, Part XV 16}, pages 35--52.
  Springer, 2020.

\bibitem{vora2020pointpainting}
Sourabh Vora, Alex~H Lang, Bassam Helou, and Oscar Beijbom.
\newblock Pointpainting: Sequential fusion for 3d object detection.
\newblock In {\em Proceedings of the IEEE/CVF conference on computer vision and
  pattern recognition}, pages 4604--4612, 2020.

\bibitem{xie2020pircnn}
Liang Xie, Chao Xiang, Zhengxu Yu, Guodong Xu, Zheng Yang, Deng Cai, and
  Xiaofei He.
\newblock Pi-rcnn: An efficient multi-sensor 3d object detector with
  point-based attentive cont-conv fusion module.
\newblock In {\em Proceedings of the AAAI conference on artificial
  intelligence}, pages 12460--12467, 2020.

\bibitem{wang2021pointaugmenting}
Chunwei Wang, Chao Ma, Ming Zhu, and Xiaokang Yang.
\newblock Pointaugmenting: Cross-modal augmentation for 3d object detection.
\newblock In {\em Proceedings of the IEEE/CVF Conference on Computer Vision and
  Pattern Recognition}, pages 11794--11803, 2021.

\bibitem{li2022deepfusion}
Yingwei Li, Adams~Wei Yu, Tianjian Meng, Ben Caine, Jiquan Ngiam, Daiyi Peng,
  Junyang Shen, Yifeng Lu, Denny Zhou, Quoc~V Le, et~al.
\newblock Deepfusion: Lidar-camera deep fusion for multi-modal 3d object
  detection.
\newblock In {\em Proceedings of the IEEE/CVF Conference on Computer Vision and
  Pattern Recognition}, pages 17182--17191, 2022.

\bibitem{zheng2022pifnet}
Wenqi Zheng, Han Xie, Yunfan Chen, Jeongjin Roh, and Hyunchul Shin.
\newblock Pifnet: 3d object detection using joint image and point cloud
  features for autonomous driving.
\newblock {\em Applied Sciences}, 12(7):3686, 2022.

\bibitem{chen2017mv3d}
Xiaozhi Chen, Huimin Ma, Ji~Wan, Bo~Li, and Tian Xia.
\newblock Multi-view 3d object detection network for autonomous driving.
\newblock In {\em Proceedings of the IEEE conference on Computer Vision and
  Pattern Recognition}, pages 1907--1915, 2017.

\bibitem{wang2019frustum}
Zhixin Wang and Kui Jia.
\newblock Frustum convnet: Sliding frustums to aggregate local point-wise
  features for amodal 3d object detection.
\newblock In {\em 2019 IEEE/RSJ International Conference on Intelligent Robots
  and Systems (IROS)}, pages 1742--1749. IEEE, 2019.

\bibitem{liang2019multitask}
Ming Liang, Bin Yang, Yun Chen, Rui Hu, and Raquel Urtasun.
\newblock Multi-task multi-sensor fusion for 3d object detection.
\newblock In {\em Proceedings of the IEEE/CVF Conference on Computer Vision and
  Pattern Recognition}, pages 7345--7353, 2019.

\bibitem{yoo20203dcvf}
Jin~Hyeok Yoo, Yecheol Kim, Jisong Kim, and Jun~Won Choi.
\newblock 3d-cvf: Generating joint camera and lidar features using cross-view
  spatial feature fusion for 3d object detection.
\newblock In {\em Computer Vision--ECCV 2020: 16th European Conference,
  Glasgow, UK, August 23--28, 2020, Proceedings, Part XXVII 16}, pages
  720--736, 2020.

\bibitem{zhou2020end}
Yin Zhou, Pei Sun, Yu~Zhang, Dragomir Anguelov, Jiyang Gao, Tom Ouyang, James
  Guo, Jiquan Ngiam, and Vijay Vasudevan.
\newblock End-to-end multi-view fusion for 3d object detection in lidar point
  clouds.
\newblock In {\em Conference on Robot Learning}, pages 923--932. PMLR, 2020.

\bibitem{lraspp2019}
Andrew Howard, Mark Sandler, Grace Chu, Liang-Chieh Chen, Bo~Chen, Mingxing
  Tan, Weijun Wang, Yukun Zhu, Ruoming Pang, Vijay Vasudevan, et~al.
\newblock Searching for mobilenetv3.
\newblock In {\em Proceedings of the IEEE/CVF international conference on
  computer vision}, pages 1314--1324, 2019.

\bibitem{ye2020hvnet}
Maosheng Ye, Shuangjie Xu, and Tongyi Cao.
\newblock Hvnet: Hybrid voxel network for lidar based 3d object detection.
\newblock In {\em Proceedings of the IEEE/CVF conference on computer vision and
  pattern recognition}, pages 1631--1640, 2020.

\bibitem{yang2018pixor}
Bin Yang, Wenjie Luo, and Raquel Urtasun.
\newblock Pixor: Real-time 3d object detection from point clouds.
\newblock In {\em Proceedings of the IEEE conference on Computer Vision and
  Pattern Recognition}, pages 7652--7660, 2018.

\bibitem{yang2018ipod}
Zetong Yang, Yanan Sun, Shu Liu, Xiaoyong Shen, and Jiaya Jia.
\newblock Ipod: Intensive point-based object detector for point cloud.
\newblock {\em arXiv preprint arXiv:1812.05276}, 2018.

\bibitem{shi2019pointrcnn}
Shaoshuai Shi, Xiaogang Wang, and Hongsheng Li.
\newblock Pointrcnn: 3d object proposal generation and detection from point
  cloud.
\newblock In {\em Proceedings of the IEEE/CVF conference on computer vision and
  pattern recognition}, pages 770--779, 2019.

\bibitem{shi2020pvrcnn}
Shaoshuai Shi, Chaoxu Guo, Li~Jiang, Zhe Wang, Jianping Shi, Xiaogang Wang, and
  Hongsheng Li.
\newblock Pv-rcnn: Point-voxel feature set abstraction for 3d object detection.
\newblock In {\em Proceedings of the IEEE/CVF Conference on Computer Vision and
  Pattern Recognition}, pages 10529--10538, 2020.

\bibitem{shi2020pointgnn}
Weijing Shi and Raj Rajkumar.
\newblock Point-gnn: Graph neural network for 3d object detection in a point
  cloud.
\newblock In {\em Proceedings of the IEEE/CVF conference on computer vision and
  pattern recognition}, pages 1711--1719, 2020.

\bibitem{yang20203dssd}
Zetong Yang, Yanan Sun, Shu Liu, and Jiaya Jia.
\newblock 3dssd: Point-based 3d single stage object detector.
\newblock In {\em Proceedings of the IEEE/CVF conference on computer vision and
  pattern recognition}, pages 11040--11048, 2020.

\bibitem{zheng2021sessd}
Wu~Zheng, Weiliang Tang, Li~Jiang, and Chi-Wing Fu.
\newblock Se-ssd: Self-ensembling single-stage object detector from point
  cloud.
\newblock In {\em Proceedings of the IEEE/CVF Conference on Computer Vision and
  Pattern Recognition}, pages 14494--14503, 2021.

\bibitem{zhang2022notequal}
Yifan Zhang, Qingyong Hu, Guoquan Xu, Yanxin Ma, Jianwei Wan, and Yulan Guo.
\newblock Not all points are equal: Learning highly efficient point-based
  detectors for 3d lidar point clouds.
\newblock In {\em Proceedings of the IEEE/CVF Conference on Computer Vision and
  Pattern Recognition}, pages 18953--18962, 2022.

\bibitem{qi2017pointnet}
Charles~R Qi, Hao Su, Kaichun Mo, and Leonidas~J Guibas.
\newblock Pointnet: Deep learning on point sets for 3d classification and
  segmentation.
\newblock In {\em Proceedings of the IEEE conference on computer vision and
  pattern recognition}, pages 652--660, 2017.

\bibitem{lang2019pointpillars}
Alex~H Lang, Sourabh Vora, Holger Caesar, Lubing Zhou, Jiong Yang, and Oscar
  Beijbom.
\newblock Pointpillars: Fast encoders for object detection from point clouds.
\newblock In {\em Proceedings of the IEEE/CVF conference on computer vision and
  pattern recognition}, pages 12697--12705, 2019.

\bibitem{deng2021voxelrcnn}
Jiajun Deng, Shaoshuai Shi, Peiwei Li, Wengang Zhou, Yanyong Zhang, and
  Houqiang Li.
\newblock Voxel r-cnn: Towards high performance voxel-based 3d object
  detection.
\newblock In {\em Proceedings of the AAAI Conference on Artificial
  Intelligence}, pages 1201--1209, 2021.

\bibitem{guan2022m3detr}
Tianrui Guan, Jun Wang, Shiyi Lan, Rohan Chandra, Zuxuan Wu, Larry Davis, and
  Dinesh Manocha.
\newblock M3detr: Multi-representation, multi-scale, mutual-relation 3d object
  detection with transformers.
\newblock In {\em Proceedings of the IEEE/CVF Winter Conference on Applications
  of Computer Vision}, pages 772--782, 2022.

\bibitem{mahmoud2023dense}
Anas Mahmoud, Jordan~SK Hu, and Steven~L Waslander.
\newblock Dense voxel fusion for 3d object detection.
\newblock In {\em Proceedings of the IEEE/CVF Winter Conference on Applications
  of Computer Vision}, pages 663--672, 2023.

\bibitem{howard2019searching}
Andrew Howard, Mark Sandler, Grace Chu, Liang-Chieh Chen, Bo~Chen, Mingxing
  Tan, Weijun Wang, Yukun Zhu, Ruoming Pang, Vijay Vasudevan, et~al.
\newblock Searching for mobilenetv3.
\newblock In {\em Proceedings of the IEEE/CVF international conference on
  computer vision}, pages 1314--1324, 2019.

\bibitem{zhou2020mvf}
Yin Zhou, Pei Sun, Yu~Zhang, Dragomir Anguelov, Jiyang Gao, Tom Ouyang, James
  Guo, Jiquan Ngiam, and Vijay Vasudevan.
\newblock End-to-end multi-view fusion for 3d object detection in lidar point
  clouds.
\newblock In {\em Conference on Robot Learning}, pages 923--932, 2020.

\bibitem{chen2022focal}
Yukang Chen, Yanwei Li, Xiangyu Zhang, Jian Sun, and Jiaya Jia.
\newblock Focal sparse convolutional networks for 3d object detection.
\newblock In {\em Proceedings of the IEEE/CVF Conference on Computer Vision and
  Pattern Recognition}, pages 5428--5437, 2022.

\bibitem{hu2022pdv}
Jordan~SK Hu, Tianshu Kuai, and Steven~L Waslander.
\newblock Point density-aware voxels for lidar 3d object detection.
\newblock In {\em Proceedings of the IEEE/CVF Conference on Computer Vision and
  Pattern Recognition}, pages 8469--8478, 2022.

\bibitem{liu2022epnet++}
Zhe Liu, Tengteng Huang, Bingling Li, Xiwu Chen, Xi~Wang, and Xiang Bai.
\newblock Epnet++: Cascade bi-directional fusion for multi-modal 3d object
  detection.
\newblock {\em IEEE Transactions on Pattern Analysis and Machine Intelligence},
  2022.

\bibitem{geiger2012kitti}
Andreas Geiger, Philip Lenz, and Raquel Urtasun.
\newblock Are we ready for autonomous driving? the kitti vision benchmark
  suite.
\newblock In {\em 2012 IEEE conference on computer vision and pattern
  recognition}, pages 3354--3361. IEEE, 2012.

\bibitem{chen2015split}
Xiaozhi Chen, Kaustav Kundu, Yukun Zhu, Andrew~G Berneshawi, Huimin Ma, Sanja
  Fidler, and Raquel Urtasun.
\newblock 3d object proposals for accurate object class detection.
\newblock {\em Advances in neural information processing systems}, 28, 2015.

\bibitem{heylen2021monocinis}
Jonas Heylen, Mark De~Wolf, Bruno Dawagne, Marc Proesmans, Luc Van~Gool, Wim
  Abbeloos, Hazem Abdelkawy, and Daniel~Olmeda Reino.
\newblock Monocinis: Camera independent monocular 3d object detection using
  instance segmentation.
\newblock In {\em Proceedings of the IEEE/CVF International Conference on
  Computer Vision}, pages 923--934, 2021.

\bibitem{openpcdet2020}
OpenPCDet~Development Team.
\newblock Openpcdet: An open-source toolbox for 3d object detection from point
  clouds.
\newblock \url{https://github.com/open-mmlab/OpenPCDet}, 2020.

\bibitem{zhang2020moca}
Wenwei Zhang, Zhe Wang, and Chen~Change Loy.
\newblock Exploring data augmentation for multi-modality 3d object detection.
\newblock {\em arXiv preprint arXiv:2012.12741}, 2020.

\end{thebibliography}

\vspace{11pt}
\begin{IEEEbiography}[{\includegraphics[width=1in,height=1.25in,clip,keepaspectratio]{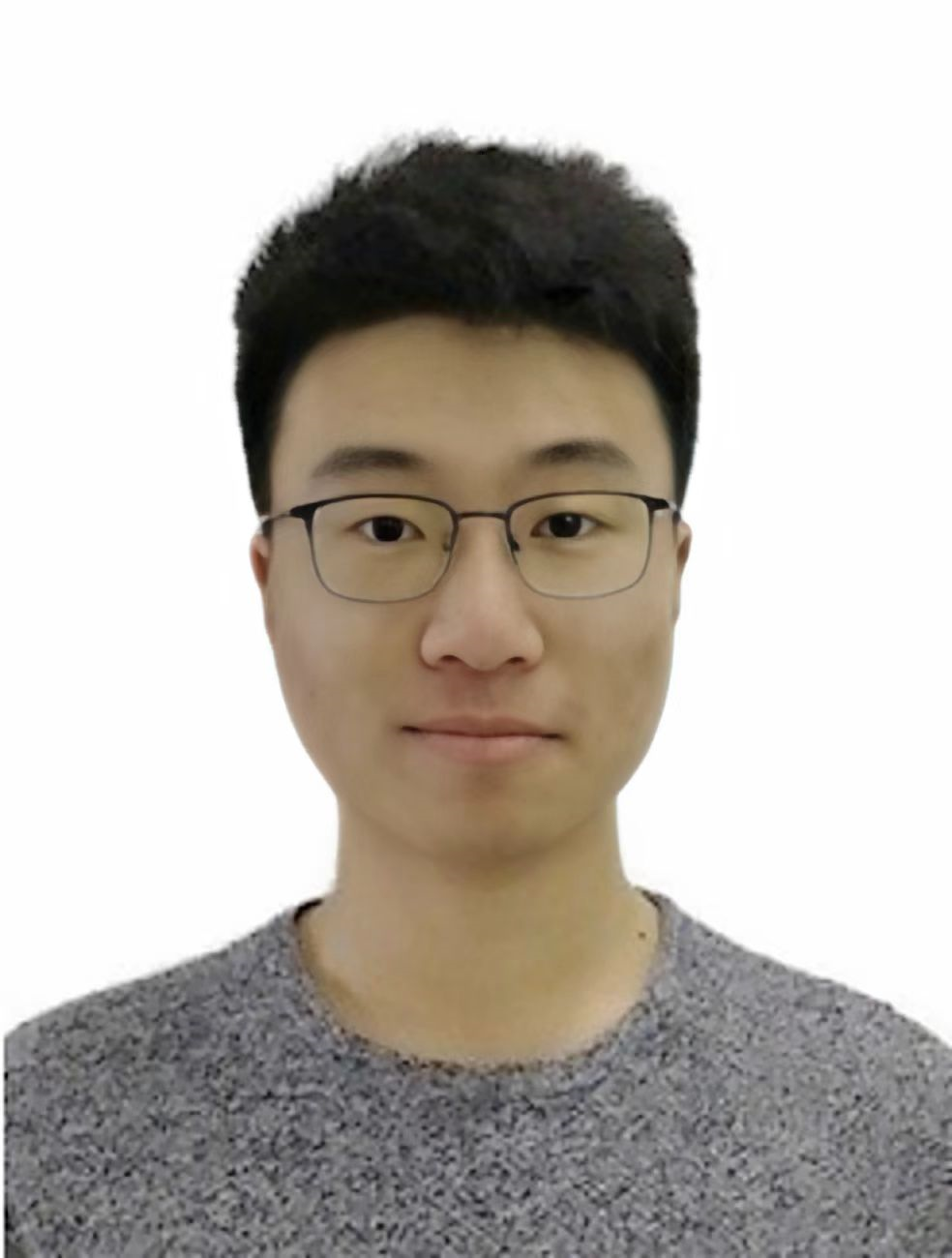}}]{Binglu Ren}
He is currently pursuing a master's degree at the School of Artificial Intelligence, Beijing University of Posts and Telecommunications, Beijing, China. His research interests include computer vision and 3D object detection.
\end{IEEEbiography}
\begin{IEEEbiography}
[{\includegraphics[width=1in,height=1.25in,clip,keepaspectratio]{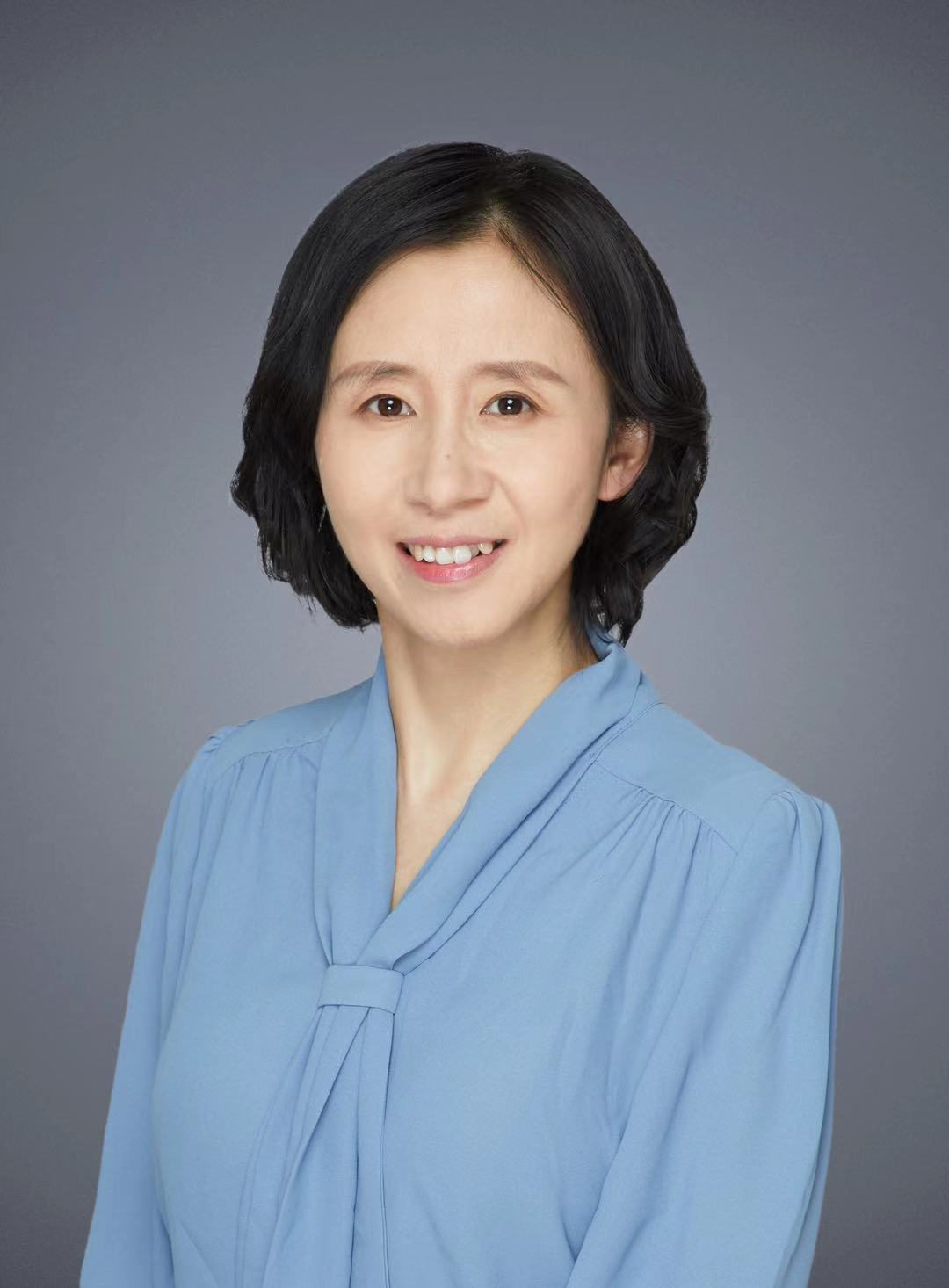}}]{Jianqin Yin}
(Member, IEEE) received the Ph.D. degree from Shandong University, Jinan, China, in 2013. She is currently a Professor at the School of Artificial Intelligence, Beijing University of Posts and Telecommunications, Beijing, China. Her research interests include service robot, pattern recognition, machine learning, and image processing.
\end{IEEEbiography}
\vfill
\end{document}